\documentclass[letterpaper, 10 pt, conference]{ieeeconf}  
\usepackage{iftex}
\usepackage{booktabs}
\ifXeTeX\usepackage[OT1]{fontenc}\fi 
 
\usepackage{amsthm}
\usepackage{amsmath}
\usepackage{amssymb}

\theoremstyle{plain}
\usepackage{algorithm}
\usepackage[noend]{algpseudocode}

\theoremstyle{definition}

\theoremstyle{remark}

\usepackage{tikz}
\usetikzlibrary{calc}
\usepackage{pgfplots}
\usepackage{placeins}
\pgfplotsset{compat=1.18}
\usepackage[font=footnotesize,labelfont=bf]{subcaption}
\usepackage[table]{xcolor}
\IEEEoverridecommandlockouts
\newcommand{\Rn}[1]{\mathbb{R}^{#1}}
\newcommand{\pvheading}[1]{\noindent\textbf{#1}}

\title{\LARGE \bf
Coupled State-Space Modelling, Control, and Policy Distillation \\ for Hybrid Rigid-Pneumatic Manipulators
\author{Alan Royce Gabriel Samuel\textsuperscript{1} and Pulkit Verma\textsuperscript{2}}
\thanks{$^{1}$ Department of Data Sciences and AI, Indian Institute of
        Technology Madras, Chennai, India
        }%
\thanks{$^{2}$ Department of Computer Science and Engineering, Indian Institute of
        Technology Madras, Chennai, India}%
\thanks{Corresponding author email: bs22b001@smail.iitm.ac.in}
}

\begin{document}
\maketitle
\thispagestyle{empty}
\pagestyle{empty}

\begin{abstract}
Hybrid manipulators combine motorized rigid joints with pressure-actuated origami segments. Published arms of this kind are controlled with decoupled per-DOF loops, and the cost of this approximation has not been quantified, because the coupled model
needed to measure it has not been built. 
This paper derives such a
model for a chain of $N$ alternating revolute joints and Kresling
origami segments, including pneumatic chamber dynamics and crease
hysteresis. 
Using the model, we measure the coupling directly and
show that its strength varies joint by joint, and that decoupled control loses
precisely on the strongly coupled joints while remaining competitive
on the one nearly decoupled joint.
Coupled
model-based controllers track $2.5\times$ tighter than a decoupled
PID baseline at lower torque. However, the model predictive
controller (MPC) is too slow for real time, and model-free
reinforcement learning stalls far below acceptable success rates on a
strict settling metric. We therefore distill the MPC into a small
neural policy with behavior cloning and DAgger. The distilled policy
settles $93$--$94\%$ of goals with zero collisions, within a few
points of its teacher, and runs inside the 5\,ms control step where
the MPC does not. Where the teacher itself fails, we trace the
failure to a limit cycle with the bellows' lightly damped mode, and
we remove it by selecting goal postures holdable at low pressure.
\end{abstract}

\section{Introduction}

\begin{figure*}[t]
  \centering
  \includegraphics[width=0.8\textwidth]{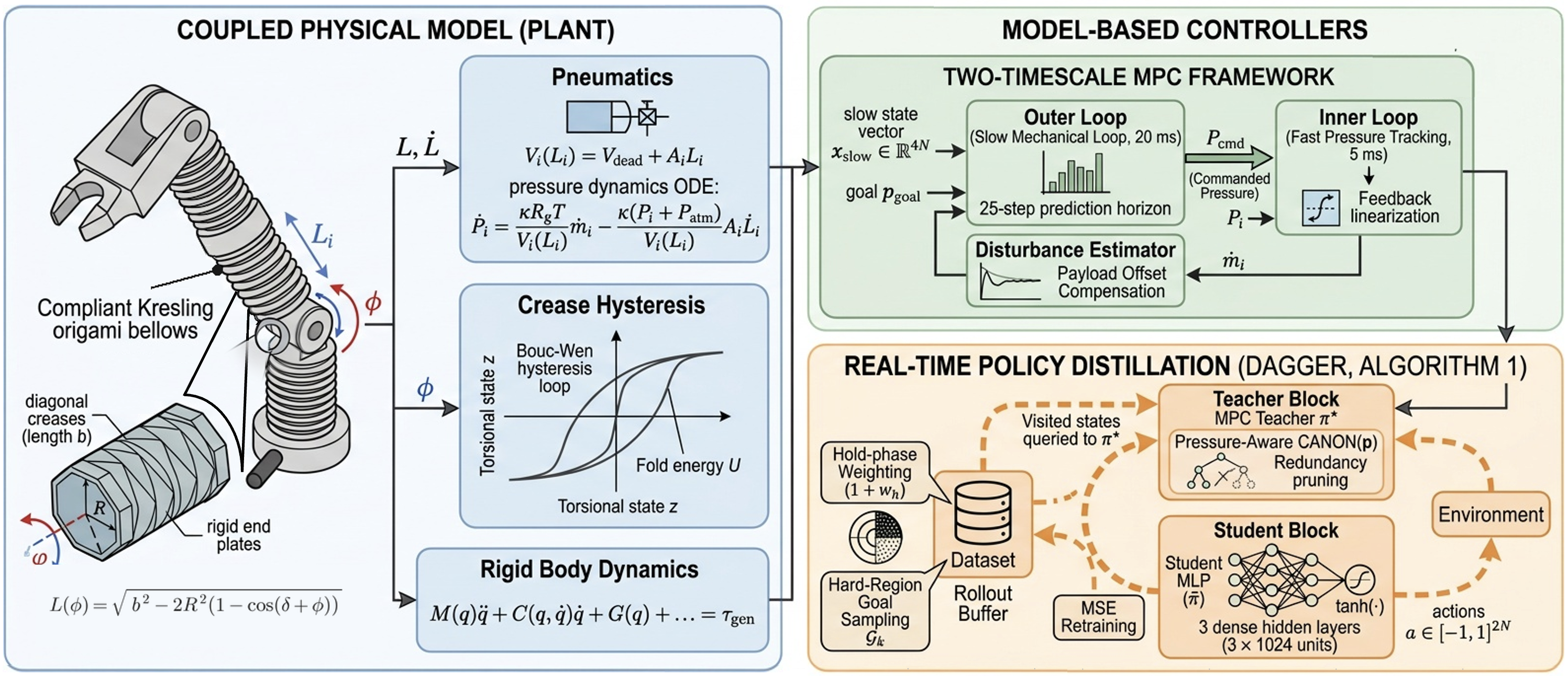}
  \caption{\textbf{Overview:}  the coupled model, the controllers it supports, and the MPC distilled into a real-time student.}
  \label{fig:overview}
\end{figure*}

How should one control a robot arm whose actuators obey different physics?
Hybrid manipulators pair two actuator classes in one
kinematic chain. Motorized revolute joints provide precise rotation,
and pneumatic origami segments provide lightweight linear extension,
with recent arms carrying payloads above 1\,kg~\cite{oh2025hybrid}. The
published answer is to pretend the question does not arise. Every such
arm is controlled with decoupled per-DOF loops, one single-input
single-output controller per joint and per chamber, with all couplings
treated as disturbance. This is an approximation, and nobody has
quantified its cost, because the coupled model that would let one
measure it has not been built. This paper builds that model and
measures what it buys (Fig.~\ref{fig:overview}).

The decoupled approximation is not obviously safe, because the two
actuator classes interact through the mechanics. Bend the arm at its
pitch joint and rotate the base, and the yaw motor drags every
bellows above it.
In the same way, a bellows that extends shifts mass
that every joint below it must carry. On our model, these
interactions are the same order as the terms a decoupled design
keeps. The normalized inertial coupling between bellows has a median
of 0.71 and never falls below 0.49 over collision-free configurations
(\S\ref{sec:experiments}). The structure of the coupling
matters as much as its size. The pitch angle alone switches the
joint-to-bellows terms on and off, and this predicts, correctly, the
one joint where decoupled control remains competitive.

Why has the coupled model not been built? First, the two
actuator classes obey different mechanics. 
A motor joint is close to an ideal torque source, so one commands a torque and gets it. 
The pressure of a pneumatic chamber, by contrast, is a dynamic state with its own differential equation.
Hence, a rigid joint carries two states, while a pneumatic segment carries three, or four once crease hysteresis is included. Second, the generalized coordinates mix units. 
The derivation itself is not the obstacle, since Lagrangian mechanics handles mixed revolute and prismatic coordinates routinely; the obstacle is every cost, tolerance and solver step built on top of it.

We derive the coupled model for a chain of $N$ alternating motorized
revolute joints and pressure-actuated Kresling origami segments. The
model joins multibody dynamics, polytropic chamber pressure dynamics,
and Bouc--Wen crease hysteresis in one state space
(\S\ref{sec:model}). Against this model we design and compare
four controllers, namely decoupled PID, LQR, a two-timescale model
predictive controller (MPC), and model-free reinforcement learning
(RL) (\S\ref{sec:control}). The comparison exposes a practical
gap. MPC gives the best combination of tracking and safety, but its
solve time already exceeds the control period at $N{=}3$. Model-free
RL is fast at deployment, but it fails to reach acceptable success
rates even after a careful redesign of the learning problem. We close
this gap by distilling the MPC into a small neural policy with DAgger
\cite{ross2011dagger} (\S\ref{sec:learning} and~\S\ref{sec:ceiling}).

The key contributions of this work are summarized below.
\begin{itemize}
\item \textbf{Coupled model.} We derive a coupled state-space model of
a hybrid rigid-pneumatic manipulator with $6N$ states, covering joint,
extension, pressure, and crease-hysteresis dynamics, calibrated
against published measurements \cite{hong2025modelbased}.
\item \textbf{Quantified coupling.} We measure the inertial coupling
directly on the model and show that it predicts where decoupled
control fails. Coupled LQR and MPC track $2.5\times$ tighter than
decoupled PID with $3.7\times$ less torque.
\item \textbf{Real-time policy by distillation.} We distill the MPC into a neural policy with behavior cloning and DAgger. 
The policy settles within 3 points of the teacher 
with zero collisions and 
it runs inside the 5\,ms control step where the MPC
does not.
\item \textbf{The teacher's ceiling.} We trace the MPC's 
residual failures at $N{=}3$ to a limit cycle with the bellows' lightly damped mode.

Selecting low-pressure goal postures among the redundant ones
removes it and lifts the teacher from $90.1\%$ to $99.7\%$.

\end{itemize}

\section{Related Work}

Oh et al.~\cite{oh2025hybrid} build a hybrid arm with rigid rotational structures and pneumatic origami chambers, but control it with decoupled per-joint PID. Zhang et al.~\cite{zhang2024origami} present a modular origami arm with Denavit--Hartenberg kinematics and a genetic-algorithm inverse kinematics, and remain purely kinematic. Hong et al.~\cite{hong2025modelbased} demonstrate model-based control of a proprioceptive origami actuator for a single unit. Koopman-operator MPC controls an origami knee exosuit with a single actuator \cite{koopman2025exo}. 
Data-driven Koopman linearizations control soft arms without a physical model \cite{bruder2021koopman}. We derive the model analytically because the controllers below need what a fitted linearization does not give: an exact collision constraint, limits on the physical inputs, and a posture-dependent coupling measurement that explains where decoupled control fails. For continuum and soft pneumatic arms, trajectory optimization with pressure dynamics and mass-flow limits \cite{falkenhahn2014trajectory} and combined dynamics and trajectory optimization for a fluidic soft arm \cite{marchese2016dynamics} do treat the pressure as a dynamic state, but for non-hybrid morphologies. Single-unit control of origami actuators is mature and continuum-arm dynamics are well studied; to our knowledge, a multi-segment coupled state space for hybrid arms is absent.

On the learning side, RL has been applied to dynamic tasks on parallel soft robots \cite{realtime2025parallel} and combined with predictive control on a cable-driven soft arm \cite{drl2026cable}. Our model-free baseline is soft actor-critic (SAC) \cite{haarnoja2018sac} with hindsight experience replay (HER) \cite{andrychowicz2017her}. Approximating an MPC law with a function approximator is long-standing \cite{parisini1995receding}, revived with neural networks for embedded deployment \cite{karg2020efficient,pan2018agile} and with guarantees \cite{hertneck2018learning}. Plain behavior cloning fails through compounding covariate shift, which aggregating labels on the learner's own states corrects \cite{ross2010efficient,ross2011dagger}; guided policy search \cite{levine2013gps} and PLATO \cite{kahn2017plato} reach similar ends, the latter with an MPC teacher adapted toward the student. These works condition the network on the optimizer's information. Our teacher conditions on a redundant joint-space target the student never observes, under a strict settling criterion, so its labels are a function of the student's observation only after canonicalization (\S\ref{sec:learning}).
Offline RL methods such as TD3+BC \cite{fujimoto2021minimalist} learn from a static dataset without querying the expert, at the price of staying near the data. Our expert is a QP rather than a robot, so querying it at the states the student visits is cheap, and DAgger uses exactly that.

\section{Coupled State-Space Model}
\label{sec:model}

\begin{figure}[t]
  \centering
  \begin{subfigure}[b]{0.44\columnwidth}
    \includegraphics[width=\linewidth]{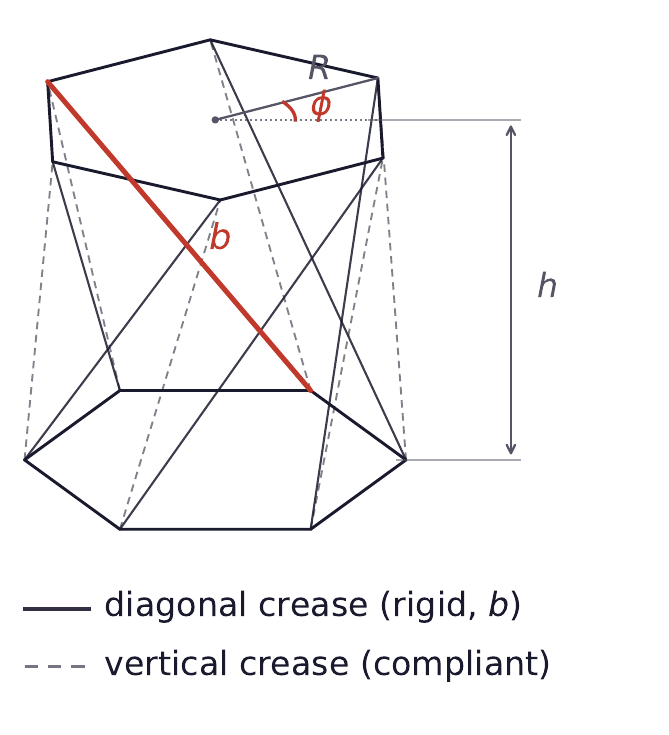}
    \caption{}
    \label{fig:kresling}
  \end{subfigure}\hfill
  \begin{subfigure}[b]{0.52\columnwidth}
    \includegraphics[width=\linewidth]{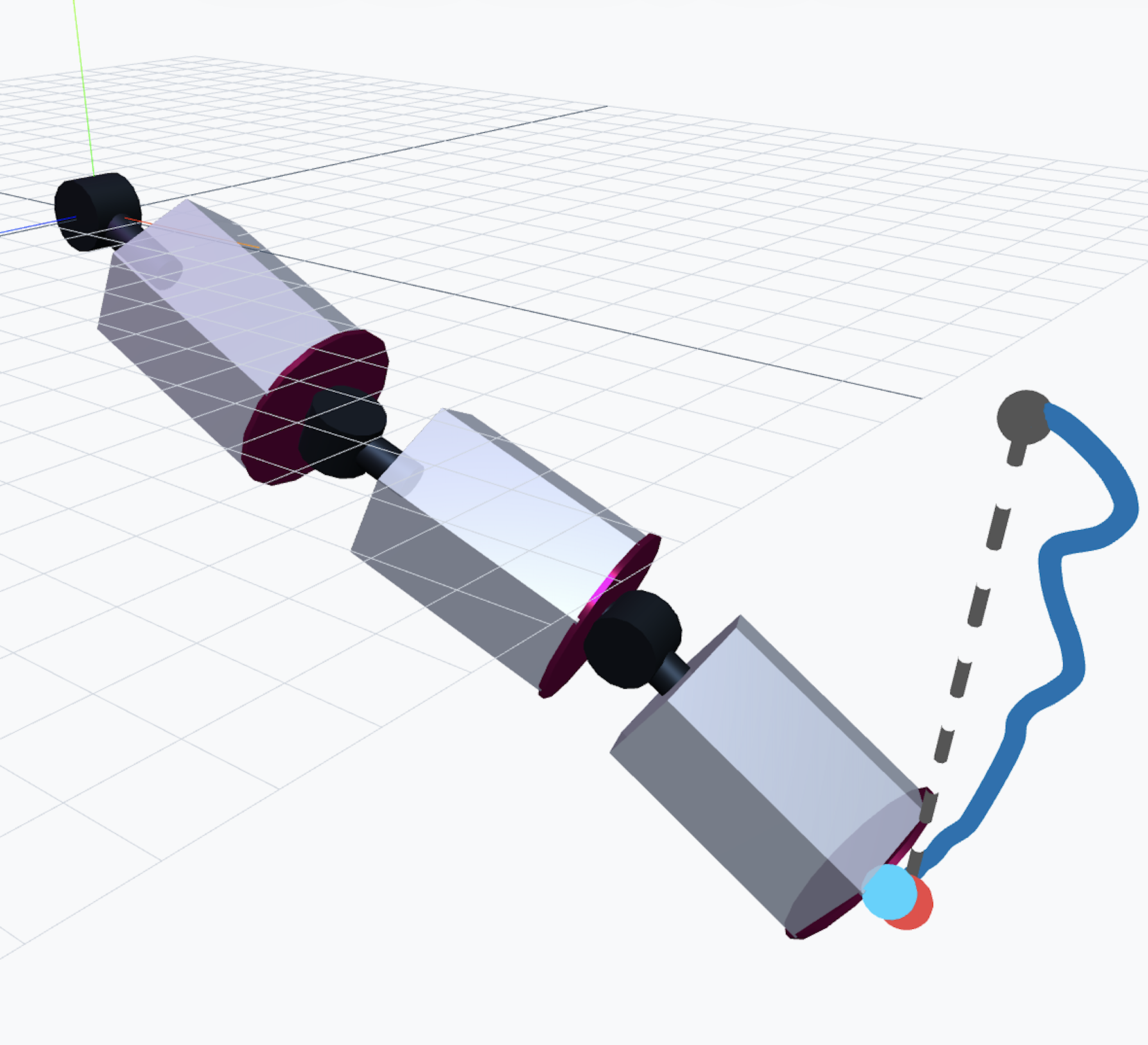}
    \caption{}
    \label{fig:live}
  \end{subfigure}
  \caption{(a) Kresling unit geometry, with rigid diagonal creases of length $b$, compliant verticals, radius $R$, twist $\phi$ and height $h$. The axial length is a geometric function of the fold angle, and the crease stiffness supplies the restoring force. Pressure extends the segment; only load or the crease spring retracts it (single-acting). (b) The assembled $N{=}3$ arm, shown as the final distilled student reaches a widened-task goal (\S\ref{sec:experiments}). The blue curve is the end-effector path and the dashed line the straight line from the start (grey dot) to the goal.}
  \label{fig:system}
\end{figure}

\pvheading{Kresling geometry.} A Kresling segment (Fig.~\ref{fig:kresling}; we call the pressurized segment the bellows) consists of two rigid polygonal plates joined by a folded side wall. Under a symmetric rigid-facet ansatz, the fold state reduces to one twist angle $\phi$. Fabrication fixes the diagonal crease length $b$, so the plate separation follows the twist:
\begin{equation}
L(\phi)=h(\phi)=\sqrt{b^2-2R^2\bigl(1-\cos(\delta+\phi)\bigr)},
\end{equation}
where $R$ is the plate radius, $h$ the plate separation, and $\delta$ a fabrication offset. The derivative $dL/d\phi$ goes to zero as $\phi$ approaches the stroke limits, so the length becomes insensitive to twist, and the crease stiffness seen along $L$ hardens there.
No regularization is needed: the simulator clips each $L_i$ to the $[54,79]$\,mm extension box at every step and every controller enforces the same box, so $d\phi/dL$ stays bounded and the stroke limits are never reached.

\pvheading{Crease hysteresis.} Real creases are path-dependent. We lump the fold energy $U$ into one torsional spring on $\phi$ and map it to an axial force by $F(L)=\frac{dU}{d\phi}\frac{d\phi}{dL}$. A Bouc--Wen state $z$ adds the hysteresis:
\begin{equation}
\tau(\phi,z)=\alpha K_\phi(\phi-\phi_0)+(1-\alpha)K_\phi z,
\end{equation}
\begin{equation}
\dot{z}=\frac{d\phi}{dL}\dot{L}\Bigl[A_{\mathrm{bw}}-|z|^{n_{\mathrm{bw}}}\bigl(\gamma_{\mathrm{bw}}+\beta_{\mathrm{bw}}\,\mathrm{sign}\bigl(\tfrac{d\phi}{dL}\dot{L}z\bigr)\bigr)\Bigr],
\end{equation}
with $K_\phi$ the crease stiffness, $\phi_0$ the rest twist, and $\alpha\in(0,1]$ the elastic fraction of the crease torque, so $1-\alpha$ is the hysteretic fraction. The Bouc--Wen parameters $\beta_{\mathrm{bw}}$ and $n_{\mathrm{bw}}$ set how fast the loop closes and how sharp its corners are, with $\gamma_{\mathrm{bw}}=\beta_{\mathrm{bw}}$ and $A_{\mathrm{bw}}=2\beta_{\mathrm{bw}}$ so $|z|$ saturates at 1\,rad.
We calibrate against the measurements of Hong et al.~\cite{hong2025modelbased}, whose reported average hysteresis loss is $4.7\%$. The fit $\alpha=0.985$, $\beta_{\mathrm{bw}}=50$, $n_{\mathrm{bw}}=1$ reproduces $4.95\%$ in simulation. The fit is window-dependent: $\alpha=0.985$ and $\alpha=0.96$ both match this target on different windows, a $2.7\times$ spread in hysteresis strength, so \S\ref{sec:experiments} sweeps $\alpha$.

\pvheading{Pneumatic chamber.} Pressure is a state with its own dynamics, and no valve command sets it instantaneously. With a polytropic lumped chamber and volume $V_i(L_i)=V_{\mathrm{dead}}+A_iL_i$, the pressure obeys
\begin{equation}
\dot{P}_i=\frac{\kappa R_g T}{V_i(L_i)}\dot{m}_i-\frac{\kappa (P_i+P_{\mathrm{atm}})}{V_i(L_i)}A_i\dot{L}_i .
\label{eq:chamber}
\end{equation}
Here $\dot m_i$ is the valve mass flow into the chamber 
, $\kappa$ the polytropic exponent, $R_gT$ the gas constant times temperature, $P_{\mathrm{atm}}$ atmospheric pressure, $A_i$ the piston area and $V_{\mathrm{dead}}$ the dead volume. The second term couples the mechanics back into the pressure. A moving piston changes the volume and hence the pressure even with the valve shut, with $\partial\dot{P}/\partial\dot{L}\approx-1.4\times10^{6}$\,Pa/m. There is also no self-restoring term, since $\partial\dot{P}_i/\partial P_i=0$ at $\dot{L}_i=0$. In open loop the chamber is a pure integrator. The constant-area volume is a control-oriented approximation: the true Kresling chamber volume varies nonlinearly with twist, and we keep the linear form so that the same $A_i$ sets the pneumatic force $A_iP_i$.

\pvheading{Multibody dynamics.} The configuration $q=[\theta_1,L_1,\dots,\theta_N,L_N]^{\top}\in\mathbb{R}^{2N}$ alternates revolute and prismatic coordinates, with joint axes alternating yaw and pitch, so at $N{=}3$ joint 2 is the only pitch joint. Forward kinematics chains per-segment homogeneous transforms, each segment a rigid housing plus the bellows extension. The task is end-effector position, so the chain is redundant by $2N-3$ for $N\geq2$. The equations of motion are
\begin{equation}
M(q)\ddot{q}+C(q,\dot{q})\dot{q}+G(q)+F_{\mathrm{elastic}}(q,z)+B_{\mathrm{damp}}\dot{q}=\tau_{\mathrm{gen}},
\label{eq:eom}
\end{equation}
where $M$ is the inertia, $C$ the Coriolis and centrifugal terms, $G$ gravity, $F_{\mathrm{elastic}}$ the crease forces including hysteresis, $B_{\mathrm{damp}}$ viscous damping, and $\tau_{\mathrm{gen}}=[\tau_i,\;A_iP_i]^{\top}$ the joint torques and bellows forces. Cross-segment coupling lives in $M(q)$, and gravity loads pass through the whole chain. Pressure enters $\tau_{\mathrm{gen}}$ algebraically and no $\dot{P}$ appears in \eqref{eq:eom}, which, with the pure-integrator pressure dynamics, licenses the two-timescale reduction of the MPC. The state $x\in\mathbb{R}^{6N}$ holds $[\theta_i,\dot{\theta}_i,L_i,\dot{L}_i,P_i,z_i]$ per segment and the input is $u=[\tau_i,\dot{m}_i]\in\mathbb{R}^{2N}$. At $N{=}3$ the rigid links are 50, 40 and 30\,mm and the extension range is $[54,79]$\,mm.

\section{Problem Setting}
\label{sec:problem}

\pvheading{Given.} The plant is the chain above, simulated with the model
of this section at a step of $dt=5$\,ms, with state
$x\in\mathbb{R}^{6N}$, input
$u=[\tau_i,\dot{m}_i]\in\mathbb{R}^{2N}$, and limits
$\tau_{\max}=5$\,N\,m, $P_{\max}=50$\,kPa and $\dot m_{\max}=5$\,g/s. A task is either a
joint-space reference trajectory or a Cartesian goal
$p_{\mathrm{goal}}\in\mathbb{R}^{3}$ drawn from a task distribution
over reachable targets; \S\ref{sec:ceiling} defines the two
distributions used at $N{=}3$.

\pvheading{Controller.} Any map from the state and the task to an
input at every step, possibly with internal state. It
may use the model, as PID, LQR, and MPC do, or only interaction data,
as the learned policies do.

\pvheading{Metrics.} For tracking, per-joint RMS error and the worst
joint's RMS torque on two fixed benchmarks, a single-target ramp and
a five-leg benchmark with sign reversals and near-limit legs. For
goal reaching, an episode starts from the home configuration, runs at
most 500 steps, and succeeds if the end effector stays inside 10\,mm
of the goal for 20 consecutive steps, called \emph{settled}
success. We also report the collision rate and per-decision wall
time. Learned-policy results average 500 episodes at an evaluation
seed disjoint from training data, with headline results confirmed
at a second seed.
A seed is a draw of 500 goals, so a rate carries a binomial 95\% interval of about $\pm2$ points, or $\pm1.5$ when two seeds are pooled; training-seed variance is measured separately in \S\ref{sec:experiments}.

\pvheading{Assumptions.} All results are in simulation, and the
simulator is the model of this section; model-based controllers use
the same model except where we perturb the plant deliberately in the
mismatch experiments. Goals are end-effector positions only, the
bellows are single-acting, and every goal is reachable by
construction.

\section{Model-Based Control}
\label{sec:control}

\pvheading{Controller suite.} We evaluate PID, LQR and MPC on the two tracking benchmarks of \S\ref{sec:problem}; model-free RL runs on the single-target benchmark. The decoupled PID baseline runs one loop per DOF, with an inner pressure loop under each bellows loop. LQR applies full-state feedback on the linearized model, and its dense gain matrix uses the cross-couplings. MPC is the two-timescale design described next. Model-free RL trains SAC \cite{haarnoja2018sac} against the simulated environment. Where a controller compensates hysteresis, it adds the Bouc--Wen crease torque $(1-\alpha)K_\phi z$, mapped to an axial force through $d\phi/dL$, to its pressure feedforward.

\pvheading{Two-timescale MPC.} Predicting the full mechanical-plus-pressure state is numerically unusable: it has a mode near $430$\,rad/s, and the largest entry of the discretised prediction matrix is about $900{,}000$ at $T_s=20$\,ms and $270{,}000$ at 5\,ms. The outer quadratic program (QP) therefore predicts only the slow state $x_{\mathrm{slow}}=[\theta_i,\dot\theta_i,L_i,\dot L_i]_{i=1}^{N}\in\mathbb{R}^{4N}$ with the commanded pressure $P_{\mathrm{cmd}}$ as input, which brings that entry down to $70$--$160$. It linearizes about the reference equilibrium every 20\,ms over a 25-step horizon, subject to torque, pressure and length boxes, with the LQR's diagonal weights minus the pressure entry: $[5\times10^{4},\,500,\,10^{5},\,100]$ on $x_{\mathrm{slow}}$ and $[500,\,10^{-4}]$ on $[\tau_i,P_{\mathrm{cmd},i}]$. An inner feedback-linearizing loop ($\lambda_P=50$\,s$^{-1}$, 5\,ms step) inverts the chamber equation \eqref{eq:chamber} to drive $P_i$ to $P_{\mathrm{cmd},i}$. An exact adjacent-bellows constraint, the collision checker's closed form, bounds the joint angle between adjacent bellows by the touching angle less a 0.10\,rad margin that covers the plant's overshoot of the linear prediction within one update. The MPC then has zero colliding steps on both benchmarks, with tracking comparable to LQR (Table~\ref{tab:rq3}). We build the QP once as a disciplined parametrized program \cite{diamond2016cvxpy,agrawal2019dpp} and re-solve it with new parameters (the \emph{cached} solve), reproducing closed-loop trajectories to $10^{-9}$ at a fraction of the cost (\S\ref{sec:experiments}).

\pvheading{Disturbance estimator.} Under sustained payloads of 0.10 and 0.25\,kg on the single-target benchmark, LQR's and MPC's final error on the loaded pitch joint grew $1.9$--$3.2\times$, while PID's integral action held its own. Neither carried integral action, so we added one shared integral-action estimator in the gravity and elastic feedforward both already compute.
It integrates the tracking error per channel, $\dot{\hat d}=-K_I\,e$ with $e=[\theta_i-\theta_{d,i},\,L_i-L_{d,i}]$, $K_I=5$ on joints and $5000$ on bellows, and subtracts $\hat d\in\mathbb{R}^{2N}$ from the feedforward $(\tau_0,P_0)$, leaving the gain and the QP untouched; under full-state measurement the offset-free disturbance observer reduces to this integrator. At 0.25\,kg it takes the pitch-joint error from $0.062$ to $0.009$\,rad for LQR and from $0.077$ to $0.001$\,rad for MPC, leaving their worst joints ($0.009$ and $0.007$\,rad) 5--7$\times$ below PID's $0.049$\,rad.

\section{Learning a Real-Time Policy}
\label{sec:learning}

MPC gives the best tracking and safety, but it is not real-time capable. It needs $34$--$118$\,ms per outer solve at $N{=}2$ against a simulation step of 5\,ms and its own cadence of 20\,ms.

\pvheading{Goal-conditioned MDP.} The MDP state is the plant state $x\in\mathbb{R}^{6N}$, and the observation $o=[x_{\mathrm{slow}},\,p_{\mathrm{ee}},\,p_{\mathrm{goal}}]$ adds the achieved and desired end-effector positions. The action $a\in[-1,1]^{2N}$ maps to $\tau_i=\tilde{\tau}_i\tau_{\max}$ and $P_{\mathrm{cmd},i}=\tfrac12(\tilde{P}_i+1)P_{\max}$, converted to valve flow by MPC's inner loop; commanding raw mass flow diverges through pressure runaway.
The bounded action keeps every actuator within its limits at any state by construction; collisions are not enforced but measured (\S\ref{sec:experiments}). Each reset draws a goal as the forward kinematics of a uniform collision-free configuration with lengths above rest. 
The per-step reward is
\begin{equation}
\begin{aligned}
r={}&-\frac{\|p_{\mathrm{ee}}-p_{\mathrm{goal}}\|}{s}-0.01\,\|a\|^{2}-\min\bigl(10\,\Phi(q),20\bigr)\\
&+10\,[\text{settled}]-100\,[\text{collision or divergence}],
\end{aligned}
\end{equation}
with $s$ the arm's maximum reach, $\Phi(q)$ the smooth collision potential of the inverse-kinematics solver with its 0.15\,rad safety margin, and the last two events ending the episode. Episodes follow the protocol of \S\ref{sec:problem}.

\pvheading{Settled success metric.} 
Under this metric, model-free RL with task-space goals and hindsight experience replay \cite{andrychowicz2017her} reaches $74.2\%$ at $N{=}1$ but only $46.8\%$ at $N{=}2$. Rigid-plant ablations separate the two costs. Removing the pressure dynamics alone lifts $N{=}2$ from $46.8\%$ to $85.0\%$, and multi-segment coordination accounts for the remaining gap. The pressure dynamics leave model-based benchmark tracking unchanged (the pressure-loop ablation of \S\ref{sec:experiments}) yet make the learning problem hard.

\pvheading{Why model-free RL stalls.} At $N{=}2$ on the full pneumatic plant, adding chamber pressure to the observation, three curricula, larger networks, 8M training steps, joint-space goals and a null-space-factored action all left settled success at or below $46.8\%$. A full MDP redesign on the tracking benchmark (target-based actions, slew-rate limits, control-rate decimation, a progressive curriculum) cut the worst-joint error $16\times$ but left it an order of magnitude above LQR and MPC. We take credit assignment across the pressure dynamics and coupled actuators, not the reward or the algorithm, as the bottleneck.

\pvheading{Distillation with DAgger.} MPC as a function, $\pi^*(o)=u_{\mathrm{MPC}}(x,x_d)$, is an accurate policy that costs a QP per decision. Distillation fits a cheap function to its decisions, a supervised problem free of the credit assignment that stalled model-free training, and DAgger \cite{ross2011dagger} fixes its known failure, compounding covariate shift (Algorithm~\ref{alg:distill}). We keep only settled teacher episodes (3{,}200 episodes, 331k transitions at $N{=}2$). The student is a stock SAC actor \cite{raffin2021sb3} fitted by mean squared error (MSE) on $\tanh(\mu)$, with targets clipped to $\pm0.999$ because 45\% of the lowest segment's pressure labels sit at the vent rail.
Fitting uses Adam at a learning rate of $10^{-3}$, batches of 2{,}048 and up to 100 epochs, and weights every sample already inside the 10\,mm tolerance, the hold phase that the dwell criterion scores, by $1+w_h$ (\S\ref{sec:experiments}). Each DAgger round rolls out the student for 1{,}000 episodes and labels every visited state, including states on the way into a collision; a failed QP solve makes MPC hold its previous command, which is not $\pi^*$, so we drop those labels
(under 0.1\% of labels in every round: 96 of 146k at $N{=}2$, and 193 of 321k and 31 of 253k in the first two $N{=}3$ rounds).

\begin{algorithm}[t]
\caption{MPC distillation with DAgger}
\label{alg:distill}
\small
\begin{algorithmic}[1]
\Require teacher $\pi^*$, goal sampler $\mathcal{G}$, action map $g$, rounds $K$
\Function{Canon}{$p$} \Comment{\S\ref{sec:learning} and~\S\ref{sec:ceiling}}
  \For{IK candidates $q$ in a fixed order}
    \If{$q$ converges and holds below 30\,kPa} \Return $q$ \EndIf
  \EndFor
  \State \Return lowest-pressure converged $q$, or reject $p$
\EndFunction
\State $\mathcal{D}\gets$ settled teacher rollouts on goals $p\sim\mathcal{G}$
\State label states: $a_t=g^{-1}\big(\pi^*(o_t;\textsc{Canon}(p))\big)$ \label{alg:labels}
\State $\hat\pi\gets\textsc{Fit}(\mathcal{D})$ \Comment{MSE, hold-phase weight $1{+}w_h$}
\For{$k=1,\dots,K$}
  \State $\mathcal{G}_k\gets\mathcal{G}$, or its diagnosed failing region
  \State roll out $\hat\pi$ on goals $p\sim\mathcal{G}_k$, visiting states $o_t$
  \State add $(o_t,a_t)$ labeled as in line~\ref{alg:labels} to $\mathcal{D}$, minus failed solves
  \State $\hat\pi_k\gets\textsc{Fit}(\mathcal{D})$ from scratch
  \If{settled success of $\hat\pi_k$ did not improve} \textbf{break} \EndIf
  \State $\hat\pi\gets\hat\pi_k$
\EndFor
\State \Return $\hat\pi$
\end{algorithmic}
\end{algorithm}

\pvheading{Canonicalization makes labels a function.} MPC conditions on a joint-space target $q_{\mathrm{goal}}$ while the student sees only $p_{\mathrm{goal}}\in\Rn{3}$. With $2N-3$ redundant DOF, two demonstrations of the same $p_{\mathrm{goal}}$ with different $q_{\mathrm{goal}}$ would demand different actions from identical student inputs, a multi-valued regression target. We remove this goal aliasing by setting $q_{\mathrm{goal}}=\mathrm{IK}(p_{\mathrm{goal}})$ with a deterministic resolved-rate solver started from home and with its joint-limit-centering term disabled. We constrain the solver to MPC's own feasibility box, namely joint limits at the collision margins and lengths between rest and the longest extension holdable within 80\% of the pressure limit. At $N{=}3$ the solver runs from a fixed ordered list of base-yaw starts to escape local minima, and we reject goals for which no start converges (3--8\%). The pressure-aware canonicalization of \S\ref{sec:ceiling} then picks among the converged candidates. The second detail is exact action inversion. The action map of the MDP above is invertible, so MPC's commands convert without loss into the student's coordinates. In the final $N{=}3$ student the joint angles enter the observation as $(\sin\theta,\cos\theta)$, since a wrapped angle is discontinuous on the goals behind the base that swing the yaw through $\pm\pi$.

\section{The Teacher's Ceiling at Three Segments}
\label{sec:ceiling}
Imitation tracks its teacher, so the teacher's failure rate at $N=3$ bounds everything downstream. With the $N{=}2$ canonicalization the teacher settled only 45\% of goals. A static feasibility filter admitting a goal only if its canonical posture holds within 80\% of the actuator limits restored 90.1\% on 3{,}551 attempts (0.9\% collisions); we call this the \emph{narrow task}. On a 300-goal diagnostic set with full failure states recorded, 268 settled, 27 timed out and 5 collided. The collisions are between non-adjacent segments 1 and 3, the one pair the outer QP does not constrain. The timeouts all enter the 10\,mm ball but oscillate across its boundary (longest streak 7 of 20 steps). They differ from successes in the goal: median equilibrium pressure 0.79 of the limit against 0.55, two bellows at the holdable ceiling against none, and 0.03\,rad errors on the load-bearing pitch joint.

We tested two hypotheses on 40 of these goals with a 1{,}500-step budget. Integral action (the estimator of \S\ref{sec:control}) settles 39 goals against 38 without it, but only 18 within the protocol's 500 steps against 38; 21 take 525--833 steps. Softer gains, or gating it to within 3\,cm of the goal, remove the slow approach but leave the last goal unsettled. Raising the pressure bound to the model's worst-case equilibrium value (158.6\,kPa, against the 50\,kPa the RL environments inherited) lowers success to 40\%. A hold test starting the arm at a goal's exact static equilibrium (Table~\ref{tab:hold}) shows MPC holding every goal at 30\,kPa and losing most at 50 and 70\,kPa, for every inner-loop gain, flow limit, outer cadence and bellows-velocity weight tried. Without the outer QP, a 1\,mm bellows kick rings out to 13\,mm at 50 and 70\,kPa but stays near 1\,mm at 30\,kPa. The ceiling is a limit cycle between the receding-horizon controller and the bellows' lightly damped mode, entered between 30 and 50\,kPa of holding pressure. It belongs to this controller on this plant, and the inherited 50\,kPa bound with its 0.8 margin (an effective 40\,kPa cap) had masked it.

The resolution is a choice of posture: with $2N-3$ redundant DOF, most goals have a posture that holds at low pressure. In \emph{pressure-aware canonicalization}, IK candidates from five base-yaw starts crossed with four ascending extension caps (73 to 78.5\,mm) are visited in a fixed order, and the goal takes the first that holds below 30\,kPa, else the lowest-pressure one. Admitting goals by best posture widens the admitted set from about half of raw draws to 89\% (of 341 draws, 20 fail IK and 17 have no holdable posture); we call this the \emph{widened task}. On 300 fresh goals, the median holding pressure falls from 0.55 to 0.29 of the limit, and the teacher settles 299 (99.7\%) without collisions. The one timeout is the farthest goal, whose best posture needs 0.52 of the limit, so the 30\,kPa boundary is approximate. The controller is untouched. The student pays instead: the rule folds the arm hard for near goals, where the imitation error concentrates (\S\ref{sec:experiments}).

\begin{table}[t]
\caption{Hold test at $N{=}3$: MPC regulates at a goal's static equilibrium for 500 steps. Held means never leaving 10\,mm, 3 goals per bin, over 25 controller variants. Last column: ring after a 1\,mm kick, outer QP removed.}
\label{tab:hold}
\centering\footnotesize\setlength{\tabcolsep}{3.5pt}
\rowcolors{3}{white}{gray!15}
\begin{tabular}{lcccc}
\toprule
Holding & Held, & Held, & Peak & Ring after \\
pressure & default & 25 variants & excursion (mm) & 1\,mm kick (mm) \\
\midrule
30\,kPa & \textbf{3/3} & 2/3--3/3 & $<$10 & 1.2--1.6 \\
50\,kPa & 1/3 & 1/3 (2/3 in 3) & 8--19 & 3.8--13.2 \\
70\,kPa & 1/3 & 0/3--1/3 & 13--39 & 7.6--13.4 \\
\bottomrule
\end{tabular}
\end{table}

\section{Experiments}
\label{sec:experiments}

We organize the evaluation around 6 research questions on the coupled model (RQ1 to RQ6), followed by the distillation results. We evaluate all controllers on the same benchmarks, and all learning results use the settled success metric.

\pvheading{Pressure modelling leaves benchmark tracking unchanged (RQ1).} We build three arms with an identical outer QP that differ only in the inner pressure loop, namely modeled, naive proportional-integral (PI), and ideal. The root-mean-square (RMS) bellows error on the single-target benchmark stays within $2.4\%$ across all three, while the pressure error differs $2\times$ without moving the mechanical tracking at all. A bandwidth sweep explains this. The modeled chamber rolls off near 8\,Hz, which attenuates the excitation that drives the bellows' lightly damped mode (14.35\,Hz at the home configuration), so the ideal-pressure arm resonates hardest and finite compressibility acts as a protective low-pass.

\pvheading{The pitch joint switches the coupling (RQ2).} We measure the unit-invariant ratio $M_{ij}/\sqrt{M_{ii}M_{jj}}$ directly on $M(q)$ over 4{,}000 collision-free configurations. The bellows-bellows block dominates with a median of 0.708 and never falls below 0.487, the joint-joint block has median 0.535, and the cross-class (joint-to-bellows) block has median 0.111. The pitch angle alone switches the cross-class block on: bending the chain couples the yaw joints to the bellows, while the pitch joint's own row of $M$ stays nearly diagonal. Per joint, the mean ratio is 0.523 for the yaw joints and 0.062 for the pitch joint. Coriolis terms contribute $0.118\%$ of the control authority at benchmark speed, a null result.

\pvheading{Coupled control tracks $2.5\times$ tighter (RQ3).} On the worst joint LQR tracks $2.5\times$ and MPC $2.4\times$ tighter than PID, with $3.7\times$ less torque (Table~\ref{tab:rq3}). PID wins only on the pitch joint (0.013 against 0.029 and 0.025\,rad), the one joint with mean coupling 0.062. On the five-leg benchmark the ranking holds: worst-joint RMS error is 0.076\,rad for PID against 0.044 for LQR and 0.060 for MPC, and RMS torque 4.81\,N\,m against 1.24 and 1.27, with no collisions. A decoupled LQR, whose gain ignores the cross-segment blocks of the linearization, tracks as well as the coupled one (0.023 against 0.029\,rad; equal on five legs), so the advantage over PID comes from full-state feedback and the model's gravity feedforward, not from cross terms in the gain. Robustness depends on the disturbance. Under a $+40\%$ mass mismatch LQR (0.031\,rad) and MPC (0.062\,rad) stay ahead of PID (0.066\,rad), because integral action cannot correct a changed transient. Under a sustained payload, a constant bias, PID's integral action wins until the disturbance estimator of \S\ref{sec:control} closes the gap.

\begin{table}[t]
\caption{Tracking on the single-target benchmark without the disturbance estimator (RQ3): RMS joint error per joint (yaw, pitch, yaw) and the worst joint's RMS torque}
\label{tab:rq3}
\centering
\setlength{\tabcolsep}{4pt} 
\rowcolors{2}{gray!15}{white}
\begin{tabular}{lcccc}
\toprule
Controller & $\theta_1$ (rad) & $\theta_2$ (rad) & $\theta_3$ (rad) & RMS $\tau$ (N\,m) \\
\midrule
PID (decoupled) & 0.073 & \textbf{0.013} & 0.064 & 4.83 \\
LQR & \textbf{0.020} & 0.029 & 0.017 & \textbf{1.32} \\
LQR (decoupled gain) & 0.023 & 0.021 & 0.017 & \textbf{1.32} \\
MPC & 0.031 & 0.025 & \textbf{0.014} & \textbf{1.32} \\
\bottomrule
\end{tabular}
\end{table}

\pvheading{Compensation makes tracking hysteresis-independent (RQ4).} As $\alpha$ falls from 0.985 to 0.80, the naive controller's tracking error grows $61\%$ relative to its value at 0.985 while the compensated controller's grows $7.5\%$.

\begin{figure*}[t]
  \centering
  \includegraphics[width=\textwidth]{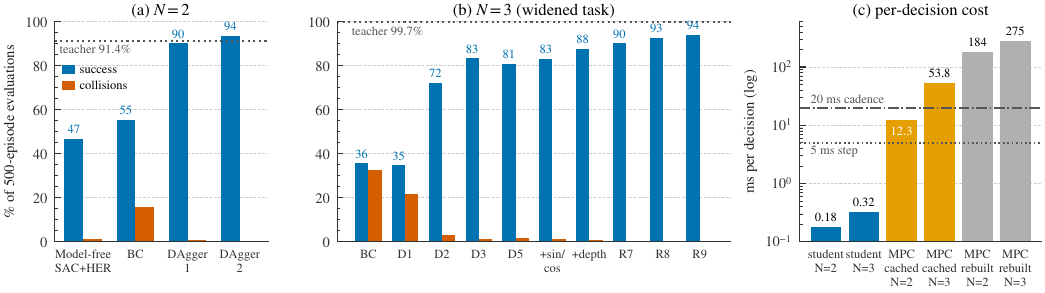}
  \caption{(a, b) Settled success and collision rate through the distillation pipeline, with 500 episodes per evaluation. The $N=2$ DAgger-1 bar pools three seeds, and DAgger-2 and all $N=3$ bars pool two. BC is behavior cloning, D$k$ is a uniform DAgger round, +sin/cos and +depth are the observation encoding and the third hidden layer, and R7--R9 are hard-region rounds (\S\ref{sec:experiments}). Panel (b) is the widened task of \S\ref{sec:ceiling}. The dotted line is the MPC teacher on the same goal protocol as the bars; on goals identical to the student's, the $N=2$ teacher settles 95.0\% (Table~\ref{tab:distill}). (c) Per-decision cost (log scale). The student is inside the 5\,ms step with a 15$\times$ margin at $N=3$; MPC exceeds the step at both $N$ and its own 20\,ms cadence at $N=3$.}
  \label{fig:progression}\label{fig:latency}
\end{figure*}

\pvheading{MPC cost grows with segment count (RQ5).} MPC's cached solve grows $4.4\times$ from $N{=}2$ to 3 (Fig.~\ref{fig:progression}(c)) and its rebuilt solve $1.5\times$
PID and LQR stay below 0.1\,ms at every $N$, and caching still leaves MPC above its cadence at $N{=}3$.

\pvheading{Model-free RL trails model-based control (RQ6).} After the MDP redesign its worst-joint tracking stays an order of magnitude behind LQR and MPC (\S\ref{sec:learning}), and settled success plateaus (Table~\ref{tab:distill}).

\pvheading{$N{=}2$: two rounds, 3 points short.} Fig.~\ref{fig:progression}(a) and Table~\ref{tab:distill} summarize 500 episodes per seed. Behavior cloning alone fails as covariate shift predicts: 0.09 RMS action error on held-out teacher states, yet $55.4\%$ success with $16\%$ collisions from a teacher that never collides, because the demonstrations contain no state near the collision boundary. One DAgger round lifts the student to $90.3\%$ pooled over three seeds, and a second removes the last collisions.
The teacher's $91.4\%$ in Table~\ref{tab:distill} is on its own demonstration goal set. On goals identical to the student's 
it settles $95.0\%$ against the student's $92.4\%$.

\begin{table}[t]
\caption{Distillation at $N{=}2$ (settled, 10\,mm, 500 episodes per seed; steps: mean steps to settle). DAgger round 2 pools two seeds at 93.2 and 94.2.}
\label{tab:distill}
\centering\footnotesize\setlength{\tabcolsep}{4pt}
\rowcolors{2}{gray!15}{white}
\begin{tabular}{lccc}
\toprule
Controller & Success & Coll. & Steps \\
\midrule
Model-free SAC + HER (plateau) & 46.8\% & 1.2\% & \\
Behavior cloning only & 55.4\% & 16.0\% & 136 \\
DAgger round 1 (3 seeds) & 90.3\% & 0.6\% & 110 \\
DAgger round 2 (2 seeds) & \textbf{93.7\%} & \textbf{0.0\%} & 106 \\
MPC teacher (demonstration goals) & 91.4\% & 0.0\% & 108 \\
MPC teacher / student, identical goals & 95.0 / 92.4\% & 0.0\% & \\
\bottomrule
\end{tabular}
\end{table}

\begin{figure}[t]
  \centering
  \includegraphics[width=0.78\columnwidth]{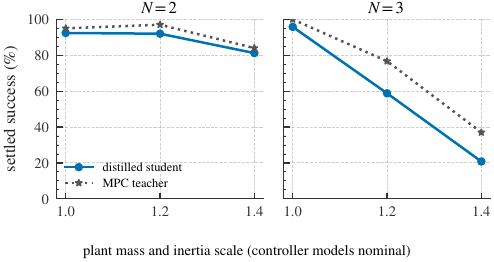}
  \caption{Mass-mismatch robustness. At $N=2$ the distilled student inherits the teacher's degradation curve. At $N=3$ both degrade steeply, the teacher through an uncorrected gravity sag and the student a further 18 points below it.}
  \label{fig:robustness}
\end{figure}

\pvheading{The student is no more robust than its teacher.} We scale the plant's masses and inertias by $+20$\% and $+40$\% while the frozen student and MPC's model stay nominal (Fig.~\ref{fig:robustness}). At $N{=}2$ the student follows the teacher's curve without collisions. At $N{=}3$ both degrade, the student more, with two collisions at $+40$\%. Every teacher miss ends 10 to 32\,mm from the goal: the nominal feedforward under-lifts the heavier bellows, and without integral action the sag remains, inside the 10\,mm tolerance at two segments but not at three. A sustained end-effector payload acts alike. At 0.25\,kg the teacher holds at $N{=}2$ (95.0\%) and collapses at $N{=}3$ (5.3\%), while the student reaches 83.0\% at $N{=}2$ and at $N{=}3$ drops to 42.4\% already at 0.10\,kg, where the teacher holds 100\%. Robustness at $N{=}3$ needs a teacher with integral action.

\pvheading{The student decides 68--170$\times$ cheaper.} On one pinned thread of an Intel i7-12700K (Fig.~\ref{fig:progression}(c)), the student's forward pass costs 0.18\,ms at $N=2$ ($[256,256]$) and 0.32\,ms at $N=3$ ($[1024{\times}3]$, 2.13M parameters), under 7\% of the 5\,ms step. The cached MPC exceeds the step at both $N$ and its own 20\,ms cadence at $N=3$, and rebuilding the QP every solve costs 5--15$\times$ more again.

\pvheading{$N{=}3$ converges below the teacher.} Distillation converged below the teacher on both tasks (Fig.~\ref{fig:progression}(b); Table~\ref{tab:n3}). On the narrow task the $N{=}2$ recipe stalled after two rounds, and two levers mattered. Widening the actor to $[1024,1024]$ took the round-3 student from about 64\% to 78.8\% while validation MSE barely moved (0.0602 to 0.0596), so aggregate MSE is a poor proxy for closed-loop behavior under near-bang-bang pressure labels. The remaining failures hovered just outside tolerance, so we weighted the 13\% of samples already inside it by $1+w_h$; $w_h=6$ gained 8 points, ending 3 below the 90.1\% teacher, as at $N{=}2$. On the widened task uniform DAgger rounds plateau at 81\% by round 5, failing on the folded postures the low-pressure rule selects for near goals (65\% on the nearest quartile against 97\% on the farthest) and on goals behind the base. The $(\sin,\cos)$ encoding and a third hidden layer recover 7 points (87.8\%), while doubling width loses 4. Validation MSE sits at a label-noise floor near 0.040 and kept epoch 24, 45 or 1 depending on the training seed, once a checkpoint that settled 27.7\%, so we select checkpoints by settled success on 100 held-out rollouts. Two to four layers then plateau within the seed spread and five is seed-sensitive (Table~\ref{tab:n3}), so we keep three. This selector alone is worth as much as the two hard-region rounds below, but did not improve the final aggregate and then plateau at 81\% by round 5. This student's failures cluster on the folded postures the low-pressure rule selects for near goals (65\% on the nearest quartile against 97\% on the farthest) and on goals behind the base. The $(\sin,\cos)$ encoding and a third hidden layer recover 7 points (87.8\% pooled), while doubling width loses 4. Early stopping on validation MSE is unreliable here, since the MSE sits at a label-noise floor near 0.040 for every architecture and kept epoch 24, 45 or 1 depending on the training seed, once keeping a checkpoint that settled 27.7\%. We therefore select checkpoints by settled success on 100 held-out rollouts. Two to four hidden layers of 1{,}024 then plateau within the seed spread while five is seed-sensitive (Table~\ref{tab:n3}), so we keep three layers. On this aggregate the selector alone is worth as much as the two hard-region rounds below combined; on the final aggregate it did not improve on the best validation-selected student (Table~\ref{tab:n3}).

\pvheading{Placed labels close half the gap.} At that plateau we diagnosed instead of tuning. A tool buckets 1{,}000 rollouts of the 87.8\% student by outcome, goal reach (distance from the base) and canonical bend, and records the \emph{action gap}: the mean distance between the student's action and the teacher's label along the student's trajectory. The failures are near-misses: 94\% end within 3\,cm, the median final distance is 16\,mm and the median in-tolerance streak is zero. They are fitting error: the action gap on failures is twice that on successes (0.60 vs 0.32), whereas compounding drift would show a small gap and a growing state error. And they are localized: 72\% success on the nearest-reach quartile, 76\% on the most-bent quartile and 79\% at base yaw above 2.2\,rad, against 94--96\% elsewhere. A localized fitting error calls for labels where the student is wrong, so a \emph{hard-region round} restricts DAgger goals to the failing half of the goal space (reach below 0.317\,m or bend above 0.6\,rad). Three gained 2.4, 2.6 and 1.4 points, where a uniform round on the same aggregate lost 3. The result is 94.2\% pooled over two seeds on the full task (1{,}000 episodes), with zero collisions there and in 1{,}000 more on the failing region (Fig.~\ref{fig:live} shows a rollout). A fourth round fell to 93.6\%, and a round confined to the innermost shell (reach below 0.30\,m, where the failures had retreated) cost 9 points overall and 12 in that shell. In this campaign label placement had an effective dose: too diffuse and near-duplicate labels dilute the aggregate, too concentrated and the skewed batch interferes globally.

Four further attempts tied or failed: teacher demonstrations on goals with reach below 0.30\,m (93.5\%), a two-policy composition switching on reach (94.0\%), a hard-region round on the aggregate with those demonstrations (93.4\%), and SAC fine-tuning from the cloned actor. Validation MSE ordered none of them. Fine-tuning, with the critic warmed up before the actor unfroze, raised the reward from $-594$ to $-51$ per episode while settled success fell to 14--15\% (5--8\% with a training-only dwell bonus): hovering near the goal minimizes the shaped per-step reward, and the fresh critic never became accurate enough to preserve a policy that cloning had placed beyond what it could evaluate. A method that keeps the actor near the data, such as TD3+BC \cite{fujimoto2021minimalist}, is the proper test, which we leave open.

\begin{table}[t]
  \caption{Three-segment arm. Settled success at seeds 1000 / 2000 (500 episodes each) with the pooled success; collision rate pooled over both seeds.}
  \label{tab:n3}
  \centering\footnotesize\setlength{\tabcolsep}{3pt}
  \resizebox{\columnwidth}{!}{%
  \rowcolors{2}{gray!15}{white}
  \begin{tabular}{lccc}
    \toprule
    Student & Success (\%) & Pooled & Coll. (\%) \\
    \midrule
    \multicolumn{4}{l}{\emph{Narrow task (feasibility-filtered goals); teacher 90.1\%, 0.9\% coll.}} \\
    Behavior cloning, $[256,256]$ & 19.4 / 16.0 & 17.7 & 44.1 \\
    DAgger round 2, $[256,256]$ & 64.8 / 63.4 & 64.1 & 3.2 \\
    DAgger round 3, $[1024,1024]$ & 79.6 / 78.0 & 78.8 & 0.9 \\
    + hold-phase weight, $w_h{=}6$ & 87.6 / 86.0 & 86.8 & 0.6 \\
    \midrule
    \multicolumn{4}{l}{\emph{Widened task (pressure-aware postures); teacher 99.7\% on 300 goals, 0 coll.}} \\
    BC, $[1024,1024]$, $w_h{=}6$ & 34.8 / 36.6 & 35.7 & 32.8 \\
    + third layer, $[1024{\times}3]$ & 90.2 / 85.4 & 87.8 & 0.6 \\
    + hard-region rounds 7--9 (final) & \textbf{95.8 / 92.6} & \textbf{94.2} & \textbf{0} \\
    same data, closed-loop sel., wd 0.01 & 93.0 / 90.2 & 91.6 & 0.2 \\
    \midrule
    \multicolumn{4}{l}{\emph{Depth ladder on the round-5 aggregate, closed-loop selection, wd 0.01}} \\
    $[1024\times 2]$ & 90.4 / 87.8 & 89.1 & 0.8 \\
    $[1024\times 3]$ & \textbf{94.0 / 89.0} & \textbf{91.5} & 0.8 \\
    $[1024\times 4]$ & 90.0 / 88.8 & 89.4 & 0.3 \\
    $[1024\times 5]$, 3-seed mean & 80.3 / 78.4 & 79.3 & 0.3 \\
    \bottomrule
  \end{tabular}}
\end{table}

\section{Conclusion}

We built the coupled state-space model that published hybrid rigid-pneumatic arms lack and used it to measure what coupling-aware control buys. The inertial coupling is large and predicts controller rankings joint by joint. Coupled LQR and MPC track tighter than decoupled PID at lower torque, and a shared disturbance estimator makes their disturbance rejection beat PID's. Because MPC is too slow for real time at three segments and model-free RL stalls on credit assignment across the pressure dynamics and coupled actuators, we distilled MPC into a small policy with DAgger, canonicalizing the teacher's joint target so its decisions are a function of the student's task-space observation, and labeling the states the student visits. The student comes within a few points of the teacher at two and three segments at a small fraction of its per-decision cost, with zero collisions on the evaluation protocol, and matches its robustness at $N{=}2$. Choosing low-pressure postures among the redundant ones avoided the teacher's own ceiling, a limit cycle with the bellows' lightly damped mode.

\pvheading{Scope.} A decoupled-gain ablation shows that the tracking
advantage over PID at benchmark speed comes from model-based
feedforward and full-state feedback, not from the cross terms. The
case for the coupling therefore rests on the coupling measurement,
the collision constraint, and higher speeds, not on
Table~\ref{tab:rq3}. Learned-policy margins come from 500-goal
draws, and we measured training-seed variance only for depth.
Nothing in the canonicalization or distillation is origami-specific
beyond the model, so the recipe should carry to other hybrid
soft-rigid chains.

In future work, we plan to pin down $\alpha$ with richer cycling
measurements. A recurrent or frame-stacked student should correct
the settle offset on folded postures, and a teacher with integral
action should extend mismatch and payload robustness to $N{=}3$.
The step that matters most is validating the distilled policy on
hardware.

\section*{Acknowledgment}
Generative AI tools were used to reword sentences and to produce the
overview illustration (Fig.~\ref{fig:overview}). All
content, results, and claims were written and verified by
the authors. We acknowledge the use of the computing resources at HPCE, IIT Madras.

\bibliographystyle{IEEEtran}
\bibliography{rl}

\end{document}